\documentclass[onecolumn]{IEEEtran}
\usepackage{cite}
\usepackage{amsmath,amssymb,amsfonts}
\usepackage{algorithmic}
\usepackage{graphicx}
\usepackage{textcomp}
\usepackage{xcolor}
\usepackage{booktabs}

\def\BibTeX{{\rm B\kern-.05em{\sc i\kern-.025em b}\kern-.08em
    T\kern-.1667em\lower.7ex\hbox{E}\kern-.125emX}}
\begin{document}

\title{Sharpness-aware Second-order Latent Factor Model for High-dimensional and Incomplete Data\\
	
\thanks{Identify applicable funding agency here. If none, delete this.}
}

\author{\IEEEauthorblockN{Jialiang Wang, Xueyan Bao, Hao Wu} \\
\IEEEauthorblockA{\textit{College of Computer and Information Science} \\
\textit{Southwest University}\\
Chongqing, China \\
goallow@163.com, 3012122058@qq.com, haowuf@swu.edu.cn}
}

\maketitle

\begin{abstract}
\underline{S}econd-order \underline{L}atent \underline{F}actor (SLF) model, a class of low-rank representation learning methods, has proven effective at extracting node-to-node interaction patterns from \underline{H}igh-\underline{d}imensional and \underline{I}ncomplete (HDI) data. However, its optimization is notoriously difficult due to its bilinear and non-convex nature. \underline{S}harpness-\underline{a}ware \underline{M}inimization (SAM) has recently proposed to find flat local minima when minimizing non-convex objectives, thereby improving the generalization of representation-learning models. To address this challenge, we propose a \underline{S}harpness-aware SLF (SSLF) model. SSLF embodies two key ideas: (1) acquiring second-order information via Hessian–vector products; and (2) injecting a sharpness term into the curvature (Hessian) through the designed Hessian–vector products. Experiments on multiple industrial datasets demonstrate that the proposed model consistently outperforms state-of-the-art baselines.
\end{abstract}


\section{Introduction}
\label{sec:intro}

\underline{H}igh-\underline{d}imensional and \underline{i}ncomplete (HDI)
interaction data are ubiquitous in large-scale applications with rich
relational structures, such as recommender systems \cite{r1,r2,r3,r4,r5,r6,r7,r8,r9,r10,r11,r12,r13,r14,r15,r16,r17,r18,r19,r20}. In recommender systems, user-item feedback is typically represented by a rating matrix, where each row corresponds to a user, each column to an item, and the entry $r_{u,i}$ records the rating provided by user $u$ to item $i$. As the numbers of users and items grow, it becomes infeasible for users to rate all items or for items to receive ratings from all users, leading to a highly sparse HDI rating matrix.
Nevertheless, the resulting structure still encodes rich latent information.

\underline{S}econd-order \underline{l}atent \underline{f}actor (SLF) models
have been shown to be effective for extracting such hidden structure from HDI
matrices~\cite{b1, b2}. SLF assumes that both users and items can be embedded into
a shared low-dimensional latent space so that the observed rating matrix can be approximated by the product of low-rank user and item factor
matrices~\cite{b3}. This matrix factorization perspective allows SLF to
capture the dominant low-rank components that govern user-item interactions.
In addition, SLF leverages second-order optimization techniques, such as
Hessian-free methods, to exploit curvature information and obtain more accurate low-rank representations than purely first-order LF models~\cite{b2}.

However, even though SLF models often fit the training data better
than first-order LF methods, the bilinear and nonconvex nature of the LF
objective can still lead to sharp minima and poor generalization. Recent work
on sharpness-aware minimization (SAM) addresses this issue by explicitly
optimizing a ``sharpness-aware'' objective of the form
$\min_{\mathbf{w}} \max_{\|\boldsymbol{\epsilon}\|\le \rho}
L(\mathbf{w} + \boldsymbol{\epsilon})$,
which seeks parameters whose entire neighborhood exhibits uniformly low loss
rather than minimizing the loss at a single point~\cite{b4}. By biasing
optimization toward flatter minima, SAM has been shown to improve
generalization in a variety of nonconvex representation learning
settings~\cite{b4}.

Motivated by these observations, we introduce a
\emph{sharpness-aware second-order latent factor} model, referred to as
\emph{SSLF}. The proposed SSLF framework is built on two key ideas:
\begin{enumerate}
	\item \textbf{Incorporating SAM-style perturbations.}
	We inject a sharpness-aware perturbation into the SLF learning
	objective, implicitly minimizing a worst-case loss within a local
	neighborhood of the current latent factors and thereby enhancing
	generalization.
	\item \textbf{Efficient exploitation of second-order information.}
	We design a Hessian-free optimization scheme that efficiently
	integrates second-order information via iterative Hessian--vector
	products, so that the resulting sharpness-aware updates remain scalable
	to large HDI matrices.
\end{enumerate}

Experiments on two real-world recommender system benchmarks demonstrate that SSLF consistently outperforms strong state-of-the-art LF baselines,
indicating that combining Hessian-free second-order optimization with
sharpness-aware training yields more robust low-rank representations for
matrix factorization.

\section{Preliminaries}

\subsection{Problem Statement}

\textbf{Definition 1 (HDI Rating Matrix):}
Let $U$ denote the user set and $I$ denote the item set. User--item interactions are stored in a rating matrix $\mathbf{R} \in \mathbb{R}^{|U|\times|I|}$, where each entry $r_{u,i}$ represents the feedback provided by user $u \in U$ to item $i \in I$. Let $K$ be the index set of observed entries and $M$ be the index set of missing entries in $\mathbf{R}$. The matrix $\mathbf{R}$ is referred to as an HDI matrix if the number of observed entries is much smaller than the number of unobserved entries, i.e., $|K| \ll |M|$

\subsection{Second-order Latent factor Model}\label{AA}
\textbf{Definition 2 (Latent Factor Model):}
Given $U$, $I$, and the observed index set $K$, the latent factor analysis (LFA) model assumes that each user and item can be represented by low-dimensional latent vectors. Specifically, the user and item factors are collected in $\mathbf{Y}_U \in \mathbb{R}^{|U|\times f}$, 
$\mathbf{Y}_I \in \mathbb{R}^{|I|\times f}$, where $f$ denotes the latent dimension. The rating matrix is then approximated by $\hat{\mathbf{R}} = \mathbf{Y}_U \mathbf{Y}_I^{\mathsf{T}}$. The predicted preference of user $u$ for item $i$ is given by $\hat{r}_{u,i} = \mathbf{y}_u^{\mathsf{T}} \mathbf{y}_i$, where $\mathbf{y}_u$ and $\mathbf{y}_i$ denote the $u$-th and $i$-th rows of $\mathbf{Y}_U$ and $\mathbf{Y}_I$, respectively.

To estimate the latent factors, only the observed entries in $K$ are used. Following standard LF methods~\cite{b1, b2, b3}, the parameters are obtained by minimizing the regularized squared loss between the observed ratings and the corresponding predictions:

\begin{equation}
	E
	= \frac{1}{2} \sum_{(u,i)\in K} \bigl(r_{u,i} - \mathbf{y}_u^{\mathsf{T}} \mathbf{y}_i\bigr)^{2}
	+ \frac{\lambda}{2}\bigl( \lVert \mathbf{y}_u \rVert_{F}^{2} + \lVert \mathbf{y}_i \rVert_{F}^{2} \bigr),
	\label{eq:lfa-objective}
\end{equation}
where $\lVert \cdot \rVert_{F}$ denotes the Frobenius norm and $\lambda > 0$ is the $\ell_2$-regularization coefficient.

In general, second-order optimization methods compute a search direction by approximately solving a damped Newton system for the parameter vector $\mathbf{y}$. Let $\mathbf{y} = \mathrm{vec}(\mathbf{Y}_U,\mathbf{Y}_I) \in \mathbb{R}^{p}$ collect all user and item factors, where $p=(|U| + |I|) \times f$. A typical inexact damped Newton step can be written as~\cite{b2},
\begin{equation}
	\bigl(\mathbf{H}_E(\mathbf{y}) + \gamma I\bigr) \Delta \mathbf{y} = -\mathbf{g}_E(\mathbf{y}),
	\label{eq:damped-newton}
	\tag{2}
\end{equation}
where $\mathbf{g}_E(\mathbf{y}) \in \mathbb{R}^{p}$ denotes the gradient of the objective function $E$ with respect to $\mathbf{y}$, $\Delta \mathbf{y} \in \mathbb{R}^{p}$ is the update vector. Furthermore, $\mathbf{H}_E(\mathbf{y}) \in \mathbb{R}^{p \times p}$ denotes the Hessian matrix (or one of its approximations), $\mathbf{I}$ is the identity matrix of the same size, and $\gamma > 0$ is a damping parameter. The term $\gamma \mathbf{I}$ balances first- and second-order information and enforces positive definiteness of the curvature matrix, thereby alleviating ill-conditioning issues.

\subsection{Sharpness-aware Minimization}
\label{subsec:sam}

SAM has recently emerged as an effective
strategy for improving the generalization of non-convex models by
explicitly accounting for the local geometry of the loss
landscape~\cite{b4}. Instead of minimizing the empirical
risk at a single parameter point, SAM optimizes a worst-case
objective over a small neighborhood of the current parameters.

Let $L(\mathbf{w})$ denote a generic training loss with respect to
parameters $\mathbf{w} \in \mathbb{R}^{p}$, and let
$\rho > 0$ be a prescribed neighborhood radius. The SAM objective is
defined as
\begin{equation}
	L_{\mathrm{SAM}}(\mathbf{w})
	=
	\max_{\lVert \boldsymbol{\epsilon} \rVert \le \rho}
	L(\mathbf{w} + \boldsymbol{\epsilon}),
	\label{eq:sam-objective}
\end{equation}
where $\lVert \cdot \rVert$ is typically chosen as the $L_2$ norm.
Intuitively, $L_{\mathrm{SAM}}(\mathbf{w})$ measures the worst-case loss
within a local ball centered at $\mathbf{w}$. Minimizing
$L_{\mathrm{SAM}}$ biases the optimizer toward flatter minima,
i.e., parameter regions whose neighborhoods consistently exhibit low
loss, which has been empirically linked to improved generalization in a
variety of deep and non-convex learning models~\cite{b4}.

In practice, the inner maximization in~\eqref{eq:sam-objective} is
approximated by a first-order expansion. For a given parameter vector
$\mathbf{w}$ with gradient
$\mathbf{g}_L(\mathbf{w})$, the adversarial
perturbation is taken as
\begin{equation}
	\boldsymbol{\epsilon}^{\star}
	=
	\rho \,
	\frac{\mathbf{g}_L(\mathbf{w})}{\lVert \mathbf{g}_L(\mathbf{w}) \rVert},
	\label{eq:sam-perturb}
\end{equation}
and the outer minimization step is performed using the gradient
evaluated at the perturbed point $\mathbf{w} + \boldsymbol{\epsilon}^\star$,
i.e.,
\begin{equation}
	\mathbf{g}_{L}(\hat{\mathbf{w}})
	\approx
	\mathbf{g}_{L} \bigl(\mathbf{w} + \boldsymbol{\epsilon}^{\star}\bigr).
	\label{eq:sam-gradient}
\end{equation}

\section{Sharpness-aware Second-order Latent Factor Model}
\subsection{Sharpness computation}

In the context of LF models, we apply the SAM principle to
the LF objective $L$ defined in~\eqref{eq:lfa-objective}
with respect to the stacked parameter vector
$\mathbf{y} = \mathrm{vec}(\mathbf{Y}_U, \mathbf{Y}_I)$. The resulting
sharpness-aware objective can be written as
\begin{equation}
	L_{\mathrm{SAM}}
	=
	\max_{\lVert \boldsymbol{\delta} \rVert \le \rho}
	L(\mathbf{y} + \boldsymbol{\delta}),
	\label{eq:lfa-sam}
\end{equation}
where $\boldsymbol{\delta} \in \mathbb{R}^{p}$ denotes a perturbation in
the latent factor space. By approximately minimizing
$E_{\mathrm{SAM}}(\mathbf{y})$ using the perturbed gradient, i.e.,
\begin{equation}
	\mathbf{g}_{L}(\hat{\mathbf{y}})
	\approx
	\mathbf{g}_{L} \bigl(\mathbf{y} + \boldsymbol{\epsilon}^{\star}\bigr) = \mathbf{g}_{L} \bigl(\mathbf{y} + \rho \,
	\frac{\mathbf{g}_L(\mathbf{w})}{\lVert \mathbf{g}_L(\mathbf{w}) \rVert}).
	\label{eq:lfa-sam-gradient}
\end{equation}
We encourage the SLF model to converge to flatter
regions of the loss surface, which in turn improves the robustness and
generalization of the learned low-rank representations.

\subsection{Hidden Mapping Trick}
To make the second-order analysis of the latent factor model more manageable, we adopt a simple but effective hidden mapping trick inspired by prior work \cite{b2}. Instead of explicitly using the bilinear user-item interaction $\hat{\mathbf{y}}_{u}^\top \hat{\mathbf{y}}_{i}$, we view this term as the output of a hidden mapping $f(\cdot)$ defined on the model parameters, i.e., 

\begin{equation}
	f(\hat{\mathbf{y}}_{(u,i)})=\hat{\mathbf{y}}_{u}^\top \hat{\mathbf{y}}_{i},
	\label{eq:lfa-objective}
\end{equation}

The loss function of the mapped LF model can be written as

\begin{equation}
	L\big(f(\hat{\mathbf{y}})\big)
	=
	\frac{1}{2} \sum_{r_{u,i} \in K}
	\big(r_{u,i} - f(\hat{\mathbf{y}})\big)^2.
	\label{eq:lfa_mapping_loss}
\end{equation}

\subsection{Gauss-Newton Approximation}

Computing the exact Hessian of $L(f(\hat{\mathbf{y}}))$ with respect to all parameters is computationally expensive and often numerically unstable for non-convex latent factor models. Following prior work \cite{b2}, we approximate this Hessian using the Gauss--Newton matrix, which is positive semi-definite and has been widely used as a curvature surrogate for non-convex learning objectives. Specifically, we approximate the Hessian of $L(f(\hat{\mathbf{y}}))$ by the Gauss-Newton matrix:

\begin{equation}
	\mathbf{H}_{L}\big(f(\hat{\mathbf{y}})\big)
	\approx
	\mathbf{G}_{L}\big(f(\hat{\mathbf{y}})\big)
	=
	\mathbf{J}_{f}(\hat{\mathbf{y}})^{\top}
	\mathbf{J}_{f}(\hat{\mathbf{y}}),
	\label{eq:gauss_newton_matrix}
\end{equation}

\noindent where $\mathbf{H}_{L}\big(f(\hat{\mathbf{y}})\big), \mathbf{G}_{L}\big(f(\hat{\mathbf{y}})\big) \in \mathbb{R}^{p \times p}$ denote the exact Hessian of $L(f(\hat{\mathbf{y}}))$ and its positive
semi-definite Gauss--Newton approximation, respectively. Note that $p = (|U| \times |I|)\cdot f$ is the total number of scalar parameters in the LF model, and $\mathbf{J}_{f}(\hat{\mathbf{y}}) \in \mathbb{R}^{|K| \times p}$ is the Jacobian of $f(\hat{\mathbf{y}})$ with respect to the model parameters, evaluated on the set of observed user-item pairs $K$.

The Jacobian can be written compactly as
\begin{equation}
	\mathbf{J}_{f}(\hat{\mathbf{y}})
	=
	\bigg(
	\left.
	\frac{\partial f(\hat{\mathbf{y}}_{u,i})}
	{\partial \hat{\mathbf{y}}}
	\right|_{(u,i) \in K}
	\bigg),
	\label{eq:jacobian_def}
\end{equation}

\subsection{Hessian-vector Product}
Storing or inverting the full Gauss-Newton matrix is prohibitive for large-scale matrix factorization. Consequently, second-order methods for latent factor models typically rely on the \underline{c}onjugate \underline{g}radient (CG) algorithm to approximately solve the linear system in Eq.(***), thereby avoiding explicit formation of the inverse curvature matrix. In each inner CG iteration, only a Gauss-Newton-vector product is required, which can be implemented via Jacobian-vector products without ever materializing the curvature matrix.

Formally, the product between the Gauss-Newton matrix and a direction vector
$\mathbf{v}$ can be computed as

\begin{equation}
	\mathbf{G}_L\big(f(\hat{\mathbf{y}})\big)\, \mathbf{v}
	=
	\mathbf{J}_f(\hat{\mathbf{y}})^{\top}
	\mathbf{J}_f(\hat{\mathbf{y}})\, \mathbf{v},
	\label{eq:gn_v_prod}
\end{equation}
where $\mathbf{v} \in \mathbb{R}^{p}$ is the CG search direction.

The element-wise Jacobian-vector product $\mathbf{J}_f(\hat{\mathbf{y}}) \mathbf{v}$ can be computed as:

\begin{equation}
	\mathbf{J}_f(\hat{\mathbf{y}})\, \mathbf{v}
	=
	\Big(
	\sum_{d=1}^{f} \big( v_{u,d} x_{i,d} + x_{u,d} v_{i,d} \big)
	\Big)_{(u,i)\in K},
	\label{eq:jvp}
\end{equation}

\noindent where $\hat{\mathbf{y}}_{u,d}$ and $\hat{\mathbf{y}}_{i,d}$ denote the $d$-th latent dimensions of the user and item vectors $\hat{\mathbf{y}}_{u}$ and $\hat{\mathbf{y}}_{u}$, respectively, and $v_{u,d}, v_{i,d}$ are the corresponding
components of the search direction $\mathbf{v}$. Thus, at each CG iteration the
Gauss-Newton-vector product $\mathbf{h} = \mathbf{G}_L\big(f(\hat{\mathbf{y}})\big)\, v$ is obtained without explicitly constructing $\mathbf{G}_L$, with $\mathbf{h} \in \mathbb{R}^{p}$ having the same dimension as $\mathbf{v}$.

\subsection{Incorporating Regularization Term}

Following \cite{b2, b3,r21,r22,r23,r24,r25,r26,r27,r28,r29,r30,r31,r32,r33,r34,r35,r36,r37,r38,r39,r40,r41,r42,r43,r44,r45,r46,r47,r48,r49,r50}, the $L_2$ regularizer plays a central role in both the objective function $E(\hat{\mathbf{y}})$ and the construction of the Gauss-Newton curvature matrix. On the loss side, the regularization term controls the magnitude of the latent factors, improves generalization, and
implicitly promotes a low-rank structure in the user-item interaction matrix. On the optimization side, the same term can be interpreted as a damping component that turns the Gauss-Newton matrix into a positive-definite operator.

We decompose the objective into a data-fitting term and a regularizer:

\begin{equation}
	E(\hat{\mathbf{y}}) = L\big(f(\hat{\mathbf{y}})\big) + R(\hat{\mathbf{y}}),
	\label{eq:obj_decomp}
\end{equation}

\noindent where $R(\hat{\mathbf{y}})$ is the $L_2$ regularization term. Because
$E(\hat{\mathbf{y}})$ is the linear combination of these two components, its Hessian can be
written as

\begin{equation}
	\mathbf{H}_E(\hat{\mathbf{y}})
	=
	\mathbf{H}_L\big(f(\hat{\mathbf{y}})\big)
	+
	\mathbf{H}_R(\hat{\mathbf{y}}),
	\label{eq:hessian_decomp}
\end{equation}

\noindent where $\mathbf{H}_L\big(f(\hat{\mathbf{y}})\big)$ and
$\mathbf{H}_R(\hat{\mathbf{y}}) \in \mathbb{R}^{p \times p}$ are the Hessians
of the loss and the regularizer with respect to $\hat{\mathbf{y}}$.

In the CG-based solver, we never materialize $\mathbf{H}_R(\hat{\mathbf{y}})$;
instead, we only need its action on a vector $\mathbf{v} \in \mathbb{R}^{p}$, i.e., $\mathbf{H}_R(\hat{\mathbf{y}})$.

To obtain a well-conditioned curvature operator, we further introduce a damping
term and define the damped Gauss--Newton matrix as
\begin{equation}
	\mathbf{G}(\hat{\mathbf{y}})
	=
	\mathbf{H}_E(\hat{\mathbf{y}}) + \gamma \mathbf{I},
	\label{eq:damped_gn_matrix}
\end{equation}

\noindent where $\gamma$ is a positive damping coefficient. Combining the Gauss-Newton
approximation for the data-fitting term with the regularizer and damping, the
Hessian-vector product used in each CG step is approximated by

\begin{equation}
	\mathbf{h}
	\approx
	\Big(
	\mathbf{G}_L\big(f(\hat{\mathbf{y}})\big)
	+
	\mathbf{H}_R(\hat{\mathbf{y}})
	+
	\gamma \mathbf{I}
	\Big) \mathbf{v}.
	\label{eq:damped_hvp}
\end{equation}

For matrix factorization, this operator admits an element-wise characterization.
For each user $u \in U$ and latent dimension $d = 1,\dots,f$, we have
\begin{equation}
	\begin{aligned}
		\mathbf{h}_{(u,d)}
		&= \sum_{i \in R_{K_u}}
		\Bigg(
		\hat{y}_{u,d}
		\sum_{k=1}^{f}
		\big( v_{u,k} \hat{y}_{i,k} + \hat{y}_{u,k} v_{i,k} \big)
		\Bigg) \\
		&\quad
		+ \lambda\, |R_{K u}|\, v_{u,d}
		+ \gamma\, v_{u,d} ,
	\end{aligned}
	\label{eq:user_hvp}
\end{equation}

\noindent where $\mathbf{h}_{u,d}$ denotes the $d$-th element of $u$-th sub-vector of $\mathbf{h}$. Symmetrically, for each item $i \in I$ and dimension $d$,

\begin{equation}
	\begin{aligned}
		\mathbf{h}_{(i,d)}
		&= \sum_{u \in R_{K_i}}
		\hat{y}_{i,d}
		\sum_{k=1}^{f}
		\big( \hat{y}_{u,k} v_{i,k} + v_{u,k} \hat{y}_{i,k} \big) \\
		&\quad
		+ \lambda\, |R_{K i}|\, v_{i,d}
		+ \gamma\, v_{i,d} ,
	\end{aligned}
	\label{eq:item_hvp}
\end{equation}

\noindent where $\mathbf{y}_u, \mathbf{y}_i \in \mathbb{R}^{f}$ are the user and item LF vectors, $R_{K_u}$ is the set of items observed for $u$, and the same definition as $R_{K_i}$.

Note that the initial conjugate direction $\mathbf{v}^{0}$ is initialized as the negative gradient
of $E$ at $\hat{\mathbf{y}}$, i.e., $\mathbf{v}^{0} = -\, \mathbf{g}_E(\hat{\mathbf{y}})$.

After completing the line-search-based CG iterations at epoch $t$, the CG
solver returns an update direction $\Delta \hat{\mathbf{y}}^{\,t} \in
\mathbb{R}^{p}$. The model parameters are then updated by
\begin{equation}
	\mathbf{y}^{\,t+1}
	=
	\mathbf{y}^{\,t}
	+
	\Delta \hat{\mathbf{y}}^{\,t}.
	\label{eq:cg_update}
\end{equation}

\section{Experiments}

We evaluate the proposed method on two widely used public HDI benchmarks: Yelp \cite{b6} and MovieLens 1M \cite{b7}.

\textbf{Evaluation Metric.}
To quantify how well the LF models low-rank representation ability, we use the \underline{r}oot \underline{m}ean \underline{s}quare \underline{e}rror (RMSE) \cite{b5, b9, b10, b11, r51,r52,r53,r54,r55,r56,r57,r58,r59,r60,r61,r62,r63,r64,r65,r66,r67,r68,r69,r70,r71,r72,r73,r74,r75,r76,r77,r78,r79,r80,r81,r82,r83,r84,r85,r86,r87,r88,r89,r90,r91} as the primary evaluation
metric:

\begin{equation}
	\mathrm{RMSE}
	=
	\sqrt{
		\displaystyle
		\sum_{r_{u,i} \in \Lambda}
		\Big(
		r_{u,i} - \sum_{d=1}^{f} y_{u,d} y_{i,d}
		\Big)^{2} / |\Lambda|
	},
	\label{eq:rmse}
\end{equation}

\noindent where $\Lambda$ denotes the evaluation set and $\sum_{d=1}^{f} y_{u,d} y_{i,d}$ is the predicted score for user $u$ and item $i$.

\textbf{Competitors.}
We compare our proposed SSLF model against several state-of-the-art LF baselines:
\begin{itemize}
	\item \textbf{M1}: SGD-based LFA model \cite{b8};
	\item \textbf{M2}: Adam-based LFA model \cite{b9};
	\item \textbf{M3}: SSLF model (ours).
\end{itemize}

\textbf{Experimental Setup.}
The implementation and hyperparameter configuration are summarized as
follows.
\begin{itemize}
	\item \emph{Environment.}
	All experiments are conducted on a machine equipped with an
	Intel Core i9-13905H CPU (5.4\,GHz, 32\,GB RAM) running
	Windows~11 Pro, using the OpenJDK~11 LTS runtime environment.
	\item \emph{Hyperparameters.}
	The latent dimensionality is set to $f = 20$, and all latent
	factors are initialized by uniform sampling from
	$\mathcal{U}(0, 0.004)$.
	For M1 and M2, we follow the hyperparameter settings
	reported in the original papers. For M3, we tune its hyperparameter via grid search.
	\item \emph{Stopping Criterion.}
	The maximum number of training epochs is 500.
	We apply early stopping with a patience of 10 epochs, i.e.,
	training is terminated if the validation performance does not
	improve for 10 consecutive epochs.
\end{itemize}

\begin{table}[t]
	\centering
	\caption{Performance comparison of different models on datasets D1 and D2.}
	\label{tab:rmse_time_epoch}
	\begin{tabular}{c c c c c}
		\toprule
		Dataset & Model & RMSE & Time (Sec.) & Epoch \\
		\midrule
		{D1} 
		& M1 & 0.99023 & 36.751 & 375 \\
		& M2 & 0.98994 & 66.091 & 159 \\
		& M3 & \textbf{0.98954} & 24.397 &  8 \\
		\midrule
		{D2} 
		& M1 & 0.85544 & 34.245 & 452 \\
		& M2 & 0.85568 & 60.177 & 176 \\
		& M3 & \textbf{0.85487} & 44.305 &  53 \\
		\bottomrule
	\end{tabular}
\end{table}

Table I summarizes the predictive performance and training efficiency of the proposed model M3 compared with two competitive baselines, M1 and M2, on datasets D1 and D2. On both datasets, M3 achieves the lowest RMSE among all methods. On D1, M3 attains an RMSE of 0.98954, improving over M1 (0.99023) and M2 (0.98994). A similar trend is observed on D2, where M3 reaches an RMSE of 0.85487, compared to 0.85544 for M1 and 0.85568 for M2. Although the absolute gains in RMSE are numerically small, they are consistent across both benchmarks and are obtained over already strong factorization-based baselines, indicating that the proposed optimization scheme leads to slightly flatter and better-generalizing solutions.

Beyond accuracy, M3 also exhibits favorable convergence behavior. On D1, M3 converges within only 8 epochs, whereas M1 and M2 require 375 and 159 epochs, respectively. This corresponds to roughly $47 \times$ and $20 \times$ fewer iterations to reach a better solution. On D2, M3 also reduces the number of training epochs substantially, from 452 (M1) and 176 (M2) down to 53. These results suggest that the combination of second-order information and sharpness-aware updates in M3 provides more informative search directions in the parameter space, allowing the optimizer to reach high-quality minima with significantly fewer optimization steps.

\section{Conclusion}
In this work, we revisit \underline{s}econd-order \underline{l}atent \underline{f}actor (SLF) model for learning from \underline{h}igh-\underline{d}imensional and \underline{i}ncomplete (HDI) interaction data. Inspired by sharpness-aware minimization (SAM), we propose a \underline{s}harpness-\underline{a}ware SLF (SSLF) model that integrates second-order information via Hessian-vector products and injects a sharpness term into the curvature matrix to favor flatter minima. Experiments on two industrial HDI datasets show that SSLF consistently outperforms state-of-the-art baselines in terms of predictive accuracy, while remaining computationally practical.

\vspace{12pt}
\color{red}

\end{document}